# The Ways of Words: The Impact of Word Choice on Information Engagement and Decision Making

Dvir, N.,[1] Friedman, E.,[1] Commuri, S.,[1] Yang, F.[1], Romano, J.[2]

[1]State University of New York at Albany, Albany, New York, NY

[2] Google

**Corresponding author:**

Nim Dvir

Department of Information Systems and Business Analytics

University at Albany, State University of New York

1400 Washington Avenue, Albany, NY 12222

ndvir@albany.edu

**Declaration of Interest statement:**

This research did not receive any specific grant from funding agencies in the public, commercial, or not-for-profit sectors.

The Ways of Words: The Impact of Word Choice on Information Engagement and Decision Making

**Abstract**

Little research has explored how *information engagement* (IE), the degree to which individuals interact with and use information in a manner that manifests cognitively, behaviorally, and affectively. This study explored the impact of phrasing, specifically word choice, on IE and decision making. Synthesizing two theoretical models, User Engagement Theory UET and Information Behavior Theory IBT, a theoretical framework illustrating the impact of and relationships among the three IE dimensions of *perception, participation,* and *perseverance* was developed and hypotheses generated. The framework was empirically validated in a large-scale user study measuring how word choice impacts the dimensions of IE. The findings provide evidence that IE differs from other forms of engagement in that it is driven and fostered by the expression of the information itself, regardless of the information system used to view, interact with, and use the information. The findings suggest that phrasing can have a significant effect on the interpretation of and interaction with digital information, indicating the importance of expression of information, in particular word choice, on decision making and IE. The research contributes to the literature by identifying methods for assessment and improvement of IE and decision making with digital text.

## 1    Introduction

The competition for viewer attention and interest continues to increase with the increasing amount of digital information available to users in many domains. Increasing reliance on computer-mediated communication via information and communication technologies (ICT) has transformed how information, specifically *digital* information, is expressed, experienced, exchanged, and employed [1]. For successful communication to occur, digital information must be transmitted through interaction, translated through interpretation, and transformed into knowledge through integration. As providing digital information in the best form and format has become crucial to attracting the target readership, it is imperative that these processes are promoted [2]. By doing so, producers and editors of text, particularly digital text, can create engagement between their end users and the information that they present to them.

## 1.1 Information Engagement

Whereas early CMC and ICT research focused on the efficiency and effectiveness of information delivery and design, current research focuses on the impact of information delivery and design on decision making and user experience (UX) [3]. In recent years, the term *engagement*, which encompasses all aspects of the user's interactions with an information system (IS), has been increasingly used to describe and measure the quality and depth of UX [4]. Describing the cognitive, behavioral, and affective (emotional) connection between a user and an IS at a point in time and possibly over time [5,6], engagement emphasizes the positive aspects of interaction with an IS and, in particular, the desire to use it longer and repeatedly [7]. For companies to increase market share and institutions to increase user awareness and knowledge, they must strive to increase user engagement with information, more specifically *information engagement* (IE), most typically in the form of digital text. Often operationalized as two-way communication between an external stakeholder (e.g., consumer, client, or citizen) and an organization (e.g., company, institution, or government agency) through various channels of correspondence, IE reflects the quality of the connection between users and information affectively, behaviorally, and cognitively, impacting the overall customer experience both online and offline and manifested as user investment and involvement [8–11].

With increasing theoretical and empirical evidence of its positive impact on users, IE is considered a significant factor in information delivery. It has become a goal—indeed, a necessity—in creating positive user interactions in various information-rich contexts and domains, including industry, government and education [12–18]. Failure to create IE has serious consequences, including lack of attention, involvement, and investment, and is associated with diminished productivity and poor decision making [19–22]. Nevertheless, many organizations fail to achieve it, resulting in failure to attract their target end users [9,17,23,24].

### 1.1.1   Information Engagement in the Digital Domain

In the digital domain, information engagement is the connection forged between an external stakeholder and an organization via the provision of information through various channels. Research into organizational communication has found that if IE is occurring, it is generally positive and desired [25,26]. In the context of government and public policy, IE is believed to indicate positive participation in the political process [27–31]. In a study of communication during the first months of the COVID-19 epidemic in the United States, IE proved important in message diffusion (transmission and retransmission), resulting in amplification of messages on social media platforms [32].

### 1.1.2   Creating and Sustaining Information Engagement

As digital text remains the most common medium of digital information transmission, failure to achieve IE with it is particularly delirious to content producers [33]. However, overcoming this failure is thwarted by a lack of guidelines for creating engaging information experiences [34,35]. Moreover, little research has focused on the development of IE, resulting in a lack of systematic approaches for its initiation, sustainment, and improvement [6,36]. Resolving this problem requires deep understanding of the IE process; the factors that influence it; and how it can be predicted and developed strategically, systematically, and computationally [6,7]. Fortunately, recent developments in computational linguistics and natural language processing (NLP) have created opportunities to explore systematic, computational, and automatic approaches to the creation, evaluation, and improvement of digital text [37].

### 1.1.3   Dimensions of Information Engagement

Research into IE in various domains has identified three primary dimensions of engagement: *perception, participation*, and *perseverance* [38]. These dimensions are significant factors in the nature of interaction with information, the level of interest in and intent for information (i.e., affect regarding, attitude toward, perception of, and evaluation of information), and the integration of information (i.e., perseverance in interacting with and application of the information). Although measurement of perception is necessary to measure engagement, it is not sufficient; *information behavior*, the totality of human behavior in relation to sources and channels of information and the nature of active and passive information seeking and use [39,40], must also be assessed. Previous research suggests that observable behavior is a reliable measure of engagement and that other aspects, such as perception and intention, are directly related to it [41,42].

#### *1.1.3.1*   **Perception**

*Perception* refers to experience with, attitude toward, relation to, emotional involvement with, interest in, and intention for information [43]. As a subjective evaluation, IE reflects a user's emotions and attitudes toward the information [44], emphasizing attracting users toward the information and thereby motivating them to interact with and use it [5,45]. Therefore, IE as perception can be conceptualized as the relationship between users and information that manifests as their level of motivation, interest, and intention toward the information, as well as an affective dimension that pertains to the quality and depth of their UX with information [45,46]. As such, IE is related to *affective involvement* and *enduring involvement,* concepts that explain why consumers, rather than becoming bored, become increasingly connected to and involved with a product over time [47,48].

### 1.1.3.2 Participation

*Participation* is the extent of participation, level of involvement, and degree of investment in information [5,45]. As it involves accessing and acquiring information, participation manifests as observable active or passive actions that can be measured by behavioral metrics. Active IE manifests by such actions as commenting on news articles [49]; clicking on, reacting to, sharing, and commenting on other users' posts [50]; signing and sharing petitions for political or policy actions [51,52]; and other forms of actively responding to digital information [10,53,54]. These behaviors not only transform into comparable performance metrics and quality indicators but also stimulate follow-up activities that replicate and expand on information across other channels and technologies [41,53,55–57].

### 1.1.3.3 Perseverance

IE is associated with a state of awareness (consciousness) and its physical manifestations that outlive the process that created them, resulting in the adoption and use of information [58]. As such, it can be described as a type of *perseverance*, a cognitive dimension manifesting as information retention, recollection, interpretation, and integration, as well as knowledge acquisition and diffusion resulting in decision making. It can be assessed by measuring these factors and their impact on decision making based on information. In the education domain, IE manifests as cognitive interaction with pedagogical materials, activities, communities, and experiences [59–65]. In the commercial domain, IE manifests as investment in an information interaction and sustained attention and interest [45,66]. IE as perseverance can also manifest as a user's level of determination to use or disposition toward information. For example, in business contexts, such as e-commerce and marketing, IE as perseverance is often used to describe the development of loyalty among engaged consumers who exhibit enhanced cognitive and emotional bonding, trust, and commitmentover a period of time, often leading to adoption and recommendation of information [7,9,67–70]

### 1.1.4    Determinants of Information Engagement

#### *1.1.4.1*    **User Determinants**

*User determinants* are personal characteristics significantly associated with IE, including age [57], gender [56,64,71,72], prior knowledge, personal relevance [73], technological knowledge and confidence [49,57,73], interests, expectations, needs, motivation, and mood [14,16,18].

#### *1.1.4.2*    **Task Determinants**

*Task determinants* relate to user goals and the activities performed to accomplish them. Previous research has highlighted type (e.g., information seeking or decision making), goal (e.g., concrete or abstract), product (e.g., format or use), frequency (e.g., discrete or repetitive), impetus (e.g., self or others), and length of time required as task determinants of IE [4]. Three stable task-related determinants are *interest, complexity,* and *difficulty*. Whereas several studies have observed a positive relationship between interest and IE [4,73–75],  there appears to be an inverse relationship between task complexity and IE. Although perceived effort appears to be a barrier to IE [6,76–78], observation of users indicates that they are attracted to tasks that are neither too difficult nor easy but at just the right level of difficulty [4,79,80]. The optimal level of difficulty, that at which the digital interaction is commensurate with level of knowledge and skill, is hypothesized to contribute to positive IE.

#### *1.1.4.3*    **System Determinants**

System determinants relate to the software and hardware features of an IS that have an impact on IE. Software determinants include the extent of interactivity provided by, the type of format(s) enabled by, and possible uses of an IS [59,63,81–83]. The primary hardware determinant is the design of the graphical interface, in particular its saliency or "visual catchiness" [54,84–86].

### *1.1.4.4* **Phrasing**

The phrasing of information, especially the choice of words, is a significant factor in the probability that information is viewed, understood, and acted upon [2,13,14,21]. Research suggests that negative phrasing (e.g., "Don't forget to bring your ID to the meeting") leads to a more passive and skeptical attitude toward the information by indicating that lack of action will lead to a problem. In contrast, positive phrasing (e.g. "Remember to bring your ID to the meeting") leads to a more active and accepting attitude toward the information by creating sense of empowerment by telling the user what to *do* rather than what to *avoid* [87]. Dvir and Gafni found that in addition to using positive phrasing, information that uses simple, clear language as opposed to complex, technical language creates more incentive to engage with the information [41]. At the same time, Agarwal and Prasad identified that use of inclusive language can foster engagement by making the information more relatable to a wider range of individuals [88]. Taken together, these findings suggest that information should be expressed as simply, clearly, and inclusively as possible, taking into account the nature of the information being transmitted.

## 1.2 **Research Gaps and Research Questions**

Recent developments in computational linguistics and natural language processing (NLP) have created opportunities to explore systematic, computational, and automatic approaches to the evaluation, creation, and improvement of digital text [89]. Despite these developments, few studies have explored the means of measuring; predicting; manipulating; and, most importantly, increasing IE systematically and computationally. Moreover, to the best of the authors' knowledge, no study has focused on the dimensions associated with *information use* rather than *technology use*, creating a significant gap in the literature. In addition, little research has examined the impact of phrasing, specifically word choice, on IE and decision making and how word choice can enhance IE. This study aimed to fill these research gaps by addressing the following research questions:

$R_1$: What are the dimensions and determinants of IE?

$R_2$: Can changes in phrasing, specifically word choice, impact IE and decision making?

## 1.3    Theoretical Underpinning

The theoretical foundation of this study is the unification of two theoretical frameworks relating to the affective, behavioral, and cognitive dimensions of IE: User Engagement Theory (UET) and Information Behavior Theory [39,40,90,91].

### 1.3.1    Information Behavior Theory

Since 1981, when he first developed the first model of Information Behavior Theory [IBT], Wilson has updated IBT over the years to accord with new findings regarding information behavior [39]. Incorporation of all Wilson's models into a unified model produced the current model of IBT to explain the way in which individuals seek, use, apply, and think about information. The model posits that (1) seeking and searching relate to information interaction and exchange; (2) information use relates to perseverance in the integration and employment of information; (3) participation is a form of information use driven by the information itself and measured by information exchange, whether passive (i.e., retrieval of information) or active (i.e., reactions to information); and (4) a positive relationship exists among selection, evaluation, and retention [39,40,90,91].

### 1.3.2    User Engagement Theory

User Engagement Theory (UET) describes the process by which *user engagement* (UE), self-directed, meaningful involvement with technological resources, is created ([45]. According to UET, UE consists of four stages—*the point of engagement, sustained engagement, disengagement,* and *potential re-engagement*—that together comprise *perseverance,* a cognitive dimension manifesting as information retention and recollection (i.e., memorization of information). The point of engagement occurs when users decide to invest in an interaction with an information source and to initiate and sustain engagement in a task. It is initiated by the aesthetic appeal and novelty of the IS interface and interest, motivation, and goal(s) in using it. A period of sustained engagement occurs when users maintain their attention and interest in the applications afforded by an IS, whereas disengagement occurs when they decide to stop using an IS or when factors in the external environment cause them to cease being engaged with it. Re-engagement occurs when users return to an IS as a result of positive experience with it. The point of engagement relates to participation, manifested as retrieval of information and the initiation of information interaction and exchange, whereas sustained engagement and re-engagement relate to perseverance, manifested in awareness, recall, and retrieval of the information from memory.

Based on UET, O'Brien and Toms developed the User Engagement Scale (UES), a measure of how information is experienced and related to perception. Analysis of the UES in various domains has shown it to be a reliable and valid measure of engagement [6,45,74,82,92,93]. Research using the UES has shown that the UE trajectory throughout use of an IS is consistent across diverse applications and that the dimensions of UE are interconnected, indicating that design of an IS must consider the entire UX experience rather than a single dimension.

## 1.4 Hypotheses

Incorporating IBT and UET, it can be hypothesized that IE is influenced by *the information itself (the expression)* and is *independent of* and *not significantly impacted by* technology, task, or user variables. Hence, variations in the phrasing, defined here as word choice, used to express the same information produce significantly different rates of participation, perception, and perseverance, and thus IE.

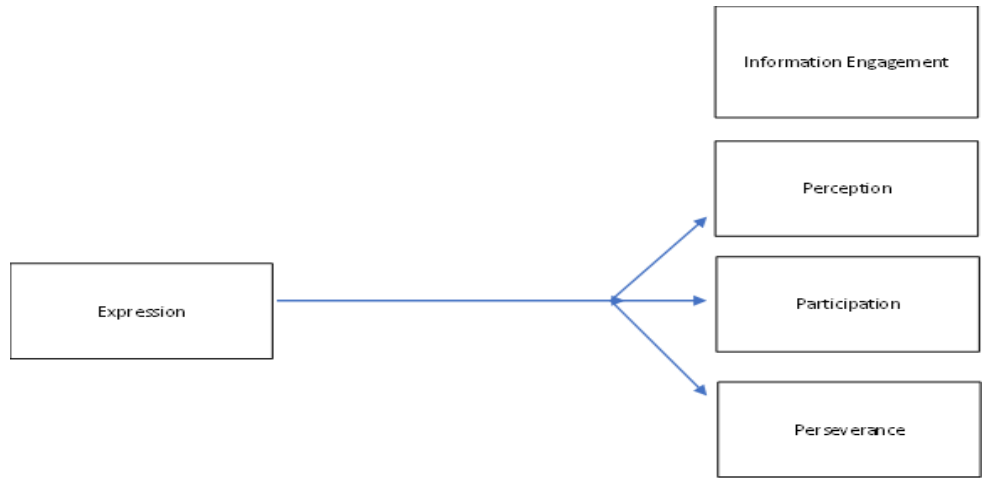

**Figure 1. Dimensions of Information Engagement**

Even synonyms, which are generally believed to express the same information, evoke different levels of IE. One word of synonym set may be more salient cognitively and affectively, and as such evoke higher or lower levels of IE. In fact, synonyms are not truly synonymous; one word may express nuances of meaning that the other does not or is attached to certain cognitive or affective events or feelings. Supporting this fact is that a *synset,* a set of synonyms, is defined as a set of cognitive synonyms *each expressing a distinct concept or meaning* [94]. For example, the words "star" and "maven" belong to the same synset because they share a meaning in WordNet ("someone who is dazzlingly skilled in any field"), yet they tend to evoke different responses in readers.

Based on these phenomena, Hypothesis 1 can be expressed as follows:

**H₁: Variations in phrasing, specifically word choice, expressing the same information produces significant differences in participation, perception, and perseverance and thus IE.**

As UET suggests positive relationships among perception, participation, and perseverance, Hypothesis 2 proposes the following:

**H₂: The IE dimensions of participation, perception, and perseverance are positively correlated.**

## 2   Methods

### 2.1   Instruments

#### 2.1.1   WordNet

The primary study instrument was a survey used to measure the participants' reactions to synonyms to assess the impact of different phrasing of the same information. To develop a survey, 100 words were chosen randomly from WordNet [95], a large lexical database of English phrases grouped into synsets. The dataset was compiled by randomly selecting the 50 synsets shown in Table 1.

**Table 1. The 50 Synsets Analyzed**

| Word1 | Word2 | Synset | Definition |
|---|---|---|---|
| abused | maltreated | abused.a.02 | Subjected to cruel treatment |
| star | maven | ace.n.03 | Someone who is dazzlingly skilled in any field |
| quick | nimble | agile.s.01 | Moving quickly and lightly |
| rich | plenteous | ample.s.02 | Affording an abundant supply |
| annoying | nettlesome | annoying.s.01 | Causing irritation or annoyance |
| art | prowess | art.n.03 | A superior skill learned by study, practice, and observation |
| gone | deceased | asleep.s.03 | Dead |
| zombie | automaton | automaton.n.01 | Someone who acts or responds in a mechanical or apathetic way |
| greedy | avaricious | avaricious.s.01 | Immoderately desirous of acquiring something, typically wealth |
| king | magnate | baron.n.03 | A very wealthy or powerful businessman |
| mother | engender | beget.v.01 | Make children |
| bubbling | belching | burp.v.01 | Expel gas from the stomach |
| fighter | belligerent | combatant.n.01 | Someone who fights or is fighting |
| computerization | cybernation | computerization.n.01 | The control of processes by computer |
| cut | shortened | cut.s.03 | With parts removed |
| lady | gentlewoman | dame.n.02 | A woman of refinement |
| death | demise | death.n.04 | The time at which life begins to end and continuing until death |
| going | departure | departure.n.01 | The act of departing |
| surrogate | deputy | deputy.n.04 | A person appointed to represent or act on behalf of others |
| find | uncovering | discovery.n.01 | The act of discovering something |

| Word1 | Word2 | Synset | Definition |
|-------|-------|--------|------------|
| dove | peacenik | dove.n.02 | Someone who prefers negotiations to armed conflict in the conduct of foreign relations |
| done | coif | dress.v.16 | To arrange attractively |
| enemy | opposition | enemy.n.02 | An armed adversary, especially a member of an opposing military force |
| foodie | epicurean | epicure.n.01 | A person devoted to refined, sensuous enjoyments, especially good food and drink |
| exile | deportee | exile.n.02 | A person who is expelled from a home or country by authority |
| lush | profuse | exuberant.s.03 | Produced or growing in extreme abundance |
| sticky | mucilaginous | gluey.s.01 | Having the sticky properties of an adhesive |
| hit | smasher | hit.n.03 | A conspicuous success |
| ill | poorly | ill.r.01 | In a poor or improper or unsatisfactory manner; not well |
| now | forthwith | immediately.r.01 | Without delay or hesitation; with no time intervening |
| reciprocation | interchange | interchange.n.02 | Mutual interaction; the activity of reciprocating or exchanging, especially information |
| natural | lifelike | lifelike.s.02 | Free from artificiality |
| just | merely | merely.r.01 | And nothing more |
| motive | need | motivation.n.01 | The psychological feature that arouses an organism to action toward a desired goal; the reason for the action that which gives purpose and direction to behavior |
| being | organism | organism.n.01 | A living thing that has or can develop the ability to act or function independently |
| pale | blanch | pale.v.01 | Turn pale, as if in fear |
| puff | gasp | pant.v.01 | Breathe noisily, as when one is exhausted |
| soul | mortal | person.n.01 | A human being |
| pirate | buccaneer | pirate.n.02 | Someone who robs at sea or plunders the land from the sea without having a commission from any sovereign nation |
| dress | primp | preen.v.03 | Dress or groom with elaborate care |
| rot | putrefaction | putrefaction.n.01 | A state of decay usually accompanied by an offensive odor |
| real | tangible | real.s.04 | Capable of being treated as fact |
| sex | gender | sex.n.04 | The properties that distinguish organisms on the basis of their reproductive roles |
| termination | ending | termination.n.05 | The act of ending something |
| right | veracious | veracious.s.02 | Precisely accurate |
| wash | lave | wash.v.02 | Cleanse one's body with soap and water |
| best | easily | well.r.03 | Indicating high probability; in all likelihood |
| better | well | well.r.11 | In a manner affording benefit or advantage |
| witch | wiccan | wiccan.n.01 | a believer in Wicca |

### 2.1.2  Qualtrics

Qualtrics[XM], a cloud-based software platform that provides a suite of tools for survey research, market research, and customer experience management, was used to collect demographic data, metadata on the technology used by the participants to complete the surveys, and survey data (i.e., responses), as well as ensure the survey was only completed once by each participant.

### 2.1.3  Participants

Prior to recruitment, the recruitment method and use of a survey was reviewed and approved by the University at Albany Institutional Review Board (IRB Study No. 22X113). Participants were recruited via a listserv sent to undergraduate students at a large research university in the United States. The inclusion criterion was active undergraduate status at the university. All participants provided informed consent before completing the online survey, after which they were asked to forward invitations to participate to their acquaintances, i.e., snowball sampling. The Qualtrics software verified that the survey was only completed once, therefore controlling for unique participants.

## 2.2 Procedure and Measurements

Data were collected using online user surveys and analyzed to assess responses to textual information (words) in terms of participation, perception and perseverance. Responses were collected using an online survey administrated through Qualtrics, Each participant was presented 7 to 10 words from the dataset in random order, with great importance placed on the randomization and control variables. Each word of the dataset was randomly presented in a random display order 805 times to randomly selected students, thereby controlling for participant characteristics. Overall, 80,500 observations were collected from 8,561 distinct participants. By balancing both known and unknown predictive factors, the survey design aimed to reduce biases, such as selection bias and allocation bias, to the greatest extent possible. Chi-square analysis of the goodness of fit of the samples revealed that the characteristic composition for each word sample was comparable to that of the overall population. Pearson chi-square analysis between demographic groups and between word samples revealed no significant differences in the composition of the demographic groups who responded to each of the words displayed.

## 2.21 Perception

To measure perception, conceptualized as the affective evaluation of information, the participants were presented with a randomized list of words and asked to evaluate it based on statements adopted from the UES [92]. The statements were presented in random order, with 4 positively formulated (e.g., "This word is easy to understand") and 4 negatively formulated (e.g., "This word is difficult to understand ).

**Table 2. Statements for Perception Evaluation Adopted From the UES**

| Code | Statement |
| --- | --- |
| EA | This word appealed to my senses. |
| EA-n | This word is not engaging. |
| FA | This word drew my attention. |
| FA-n | I wasn't focused while reading this word. |
| PU | This word was easy to understand. |
| PU-n | This word was difficult to understand. |
| RW | The experience reading this word was rewarding. |
| RW-n | The experience reading this word is not worthwhile. |

The statements were used to assess each word's sensory appeal, ability to stimulate attention, perceived usability, and reward. The participants indicated how much they agreed or disagreed with each statement using a 5-point scale that ranged from 1 (*strongly disagree*) to 5 (*strongly agree*). The responses to the negative statements were reverse-coded to maintain a unified 5-item scale. Testing for scale reliability yielded a Cronbach's alpha of 0.888.

### 2.1.4   Participation

To measure participation, operationalized as the selection and retrieval of information and behavioral reactions to it, the participants were asked to click on the words with which they would like to engage. Their responses were recorded as 1 for *engaged* and 0 for not *engaged* and the selection rate calculated.

### 2.1.5   Perseverance

To measure perseverance, operationalized as the retention of information, the participants were asked to write down the words that they remembered from a list that had been presented. Their responses were recorded as 1 for *remembered* and 0 for *not remembered* and the selection and retention rates calculated.

## 3   Results

### 3.1   Demographics

The mean age of the participants was 22.1 (SD = 1.388), with 75.3% of responses made by the age group of 17–22, 41.2% by females, and 58.8% by males. More than 95.0% reported that their English proficiency was at least "very good," with 79.0% identifying as native speakers. The majority of the observations were submitted through a laptop device (70.1%), followed by mobile devices (15.6%) and desktops (14.2%). This may be attributed to the survey being easier to read and complete on a large screen, as was reported by participants in early stages of the survey design.

### 3.2   Statistical Testing

Figures 2, 3, and 4 show the mean rates of the three IE dimensions for the 100 words examined, with participation operationalized as selection rate, perception operationalized as evaluation (average UES score as percentage of 5), and perseverance operationalized as retention rate.

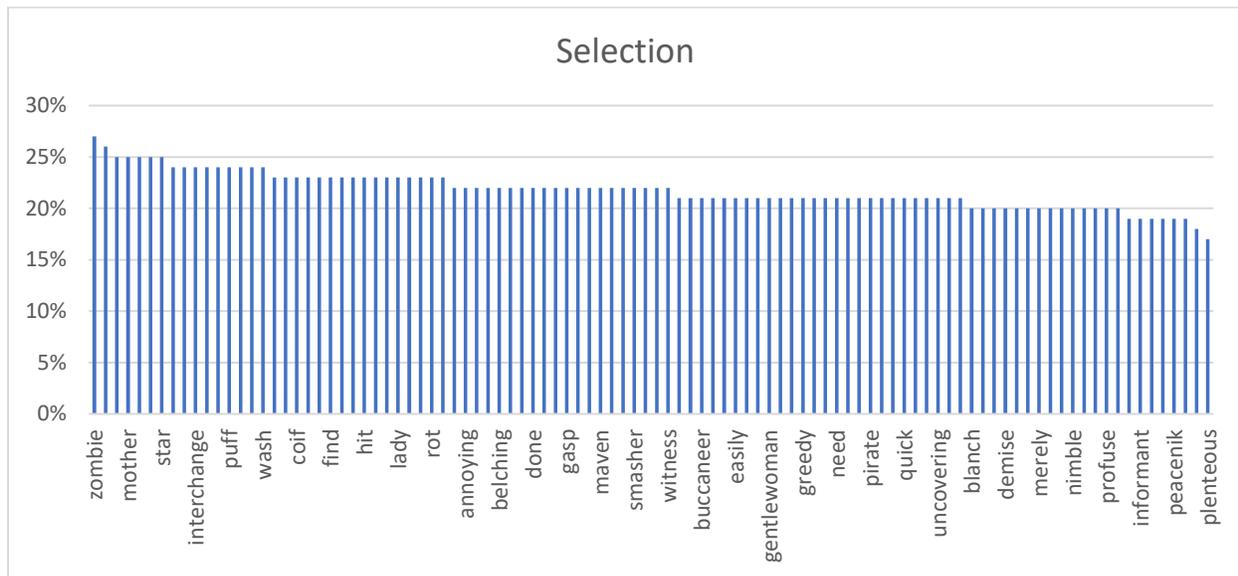

**Figure 2. Words Selection Rates**

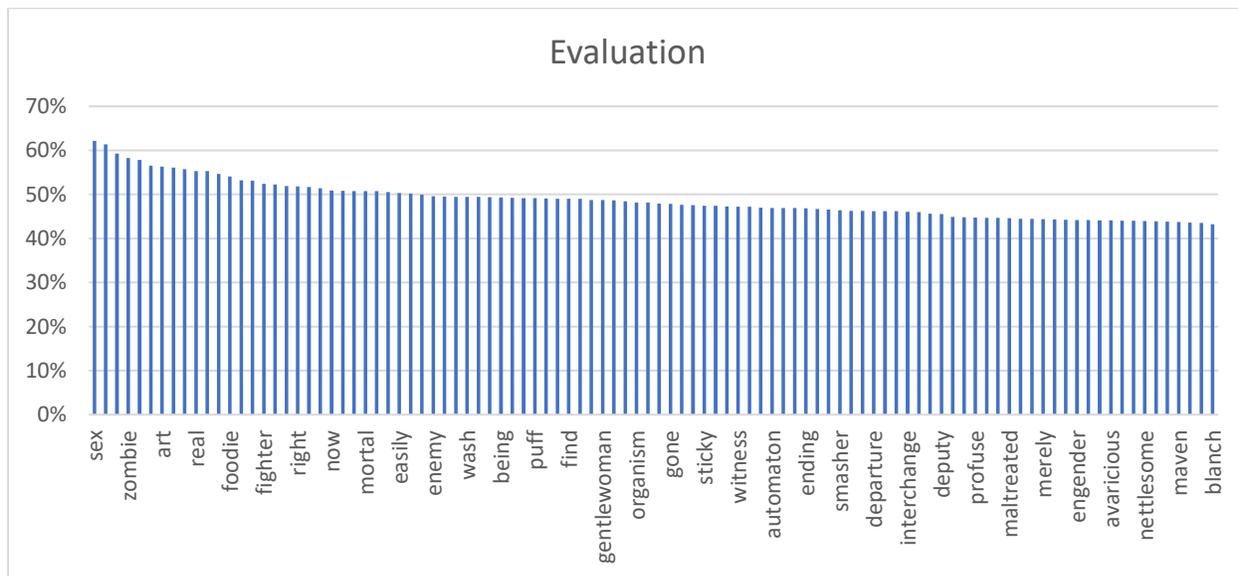

**Figure 3. Words Evaluation Rates**

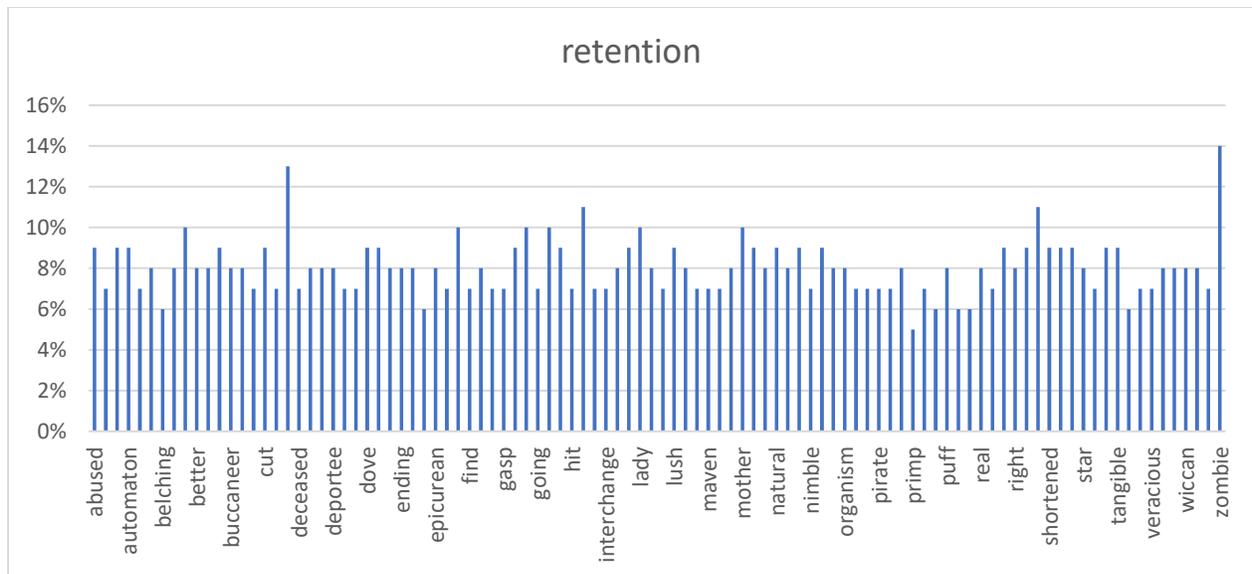

**Figure 4. Words Retention Rates**

A one-way ANOVA conducted to compare the evaluation, selection, and retention rates among words revealed significant differences among words in evaluation rate (F[99], 80400) = 48.579, p = .000, **η2**=.056), selection rates (F[99, 80400] = 1.481, p = .001, **η2**=.002), and retention rates (F[99], 80400) = 1.921, p = .001, **η2**=.002). The table below shows the results in more detail.

**Table 3. ANOVA Results for Comparison of Evaluation, Selection, and Retention Rates**

|  |  | SS | df | MS | F | Sig. | **η2** |
|---|---|---|---|---|---|---|---|
| Evaluation | Between Groups | 3497.624 | 99 | 35.330 | 48.579 | .000 | .056 |
|  | Within Groups | 58471.551 | 80400 | .727 |  |  |  |
|  | Total | 61969.175 | 80499 |  |  |  |  |
| Selection | Between Groups | 24.985 | 99 | .252 | 1.481 | .001 | .002 |
|  | Within Groups | 13702.281 | 80400 | .170 |  |  |  |
|  | Total | 13727.265 | 80499 |  |  |  |  |
| Retention | Between Groups | 14.087 | 99 | .142 | 1.921 | <.001 | .002 |
|  | Within Groups | 5956.037 | 80400 | .074 |  |  |  |
|  | Total | 5970.124 | 80499 |  |  |  |  |

Single sample t-tests ($\bar{X} \neq \mu$) were conducted to compare each word's mean for the three dimensions and to compare general population means to test for significant differences. Regarding perception, 76 words had evaluation means significantly different than the population mean, 30 of which had evaluation means significantly greater and 48 words had means significantly smaller than the population mean. Regarding participation, 17 words had selection means significantly different than the population mean, 7 of which were higher and 10 lower. Regarding perseverance, 18 words had mean retention rates significantly different than the population mean, 8 of which were higher and 10 lower.

To determine if selection, evaluation, and retention rates are impacted by word choice regardless of the meaning, pairwise comparisons for independent means ($\bar{X}_1 \neq \bar{X}_2$) were conducted to evaluate difference between synonymous words for each of the 50 synsets. By using 50 pairs of words for independent samples t-test, the evaluation aimed to prevent the possibility of a Type I error, i.e., mistakenly rejecting the null. The results revealed that 43 of the 50 synsets had significantly different IE scores ($P < 0.05$).

The charts below show the difference for selected synsets.

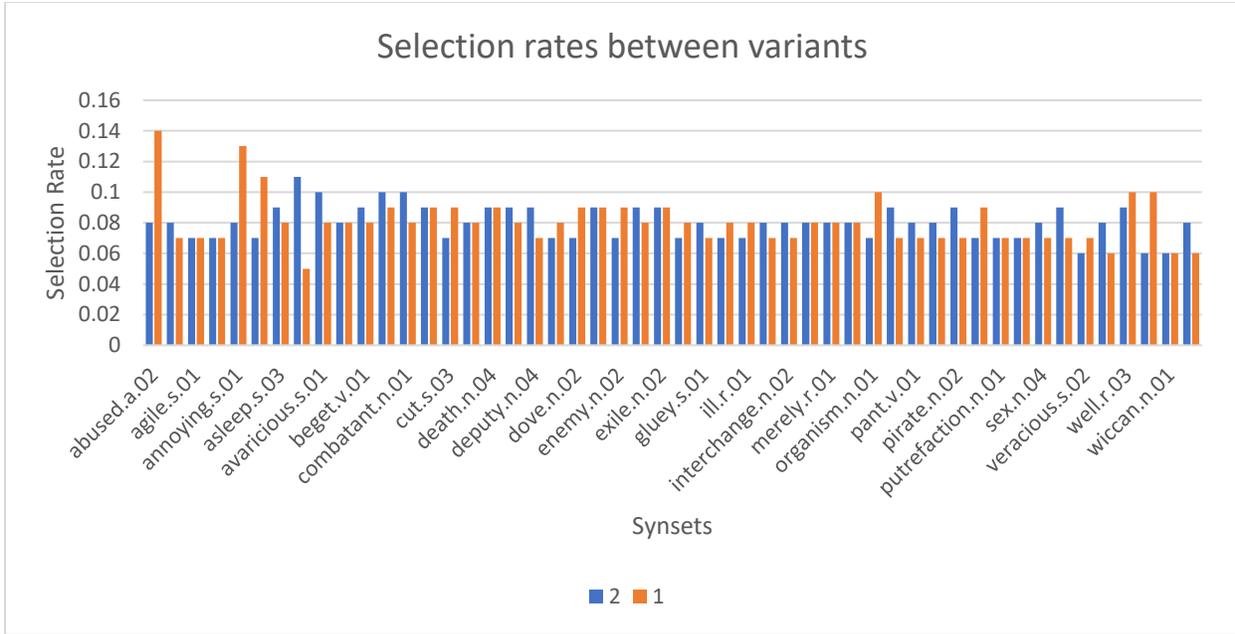

**Figure 5. Difference in selection rates between word pairs (variants in synsets)**

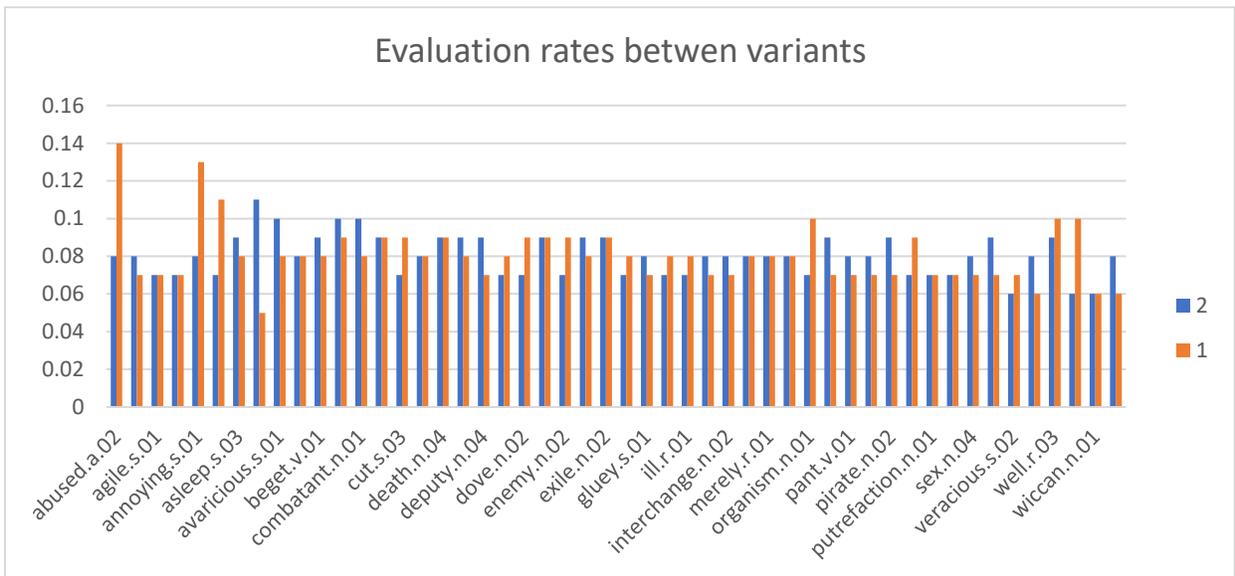

**Figure 6. Difference in evaluation rates between word pairs**

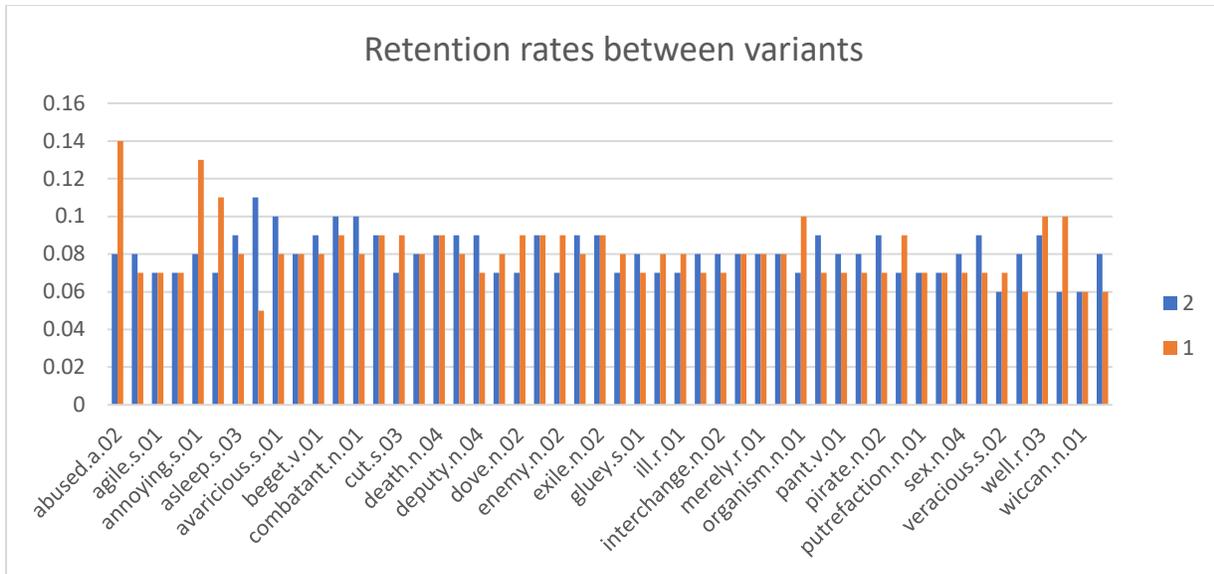

**Figure 7. Difference in retention rates between word pairs**

### 3.3   Hypothesis Testing

#### 3.3.1   Hypothesis 1

**H₁: Variations in phrasing, specifically word choice, expressing the same information produces significant differences in participation, perception, and perseverance and thus IE.**

The results indicate that some words have mean perception, selection, and retention rates significantly different from the mean rates. The results of single sample t-tests confirmed that some words are more or less engaging than others, regardless of user demographics, word order, or technology used to view text. Together these results indicate that variations in wording lead to differences in the IE dimensions, providing evidence with which to accept Hypothesis 1 and reject the null.

#### 3.3.2   Hypothesis 2

**H₂: The IE dimensions of participation, perception, and perseverance are positively correlated.**

A statistically significant positive correlation was found between a word's UES score (evaluation rate) and its selection rate, $r(100) = .64$, $p < .001$, and retention rate, $r(100) = .63$. These correlations indicate that a positive correlations exists among the three dimensions, providing evidence with which to reject the null and accept Hypothesis 2.

**Table 4. Correlations Among dimensions of 100 Words**

| | | select | | retention |
|---|---|---|---|---|
| Evaluation | Pearson Correlation | .636** | 1 | .625** |
| | Sig. (2-tailed) | <.001 | | <.001 |
| | Sum of Squares and Cross-products | .060 | .293 | .045 |
| | Covariance | .001 | .003 | .000 |
| | N | 100 | 100 | 100 |

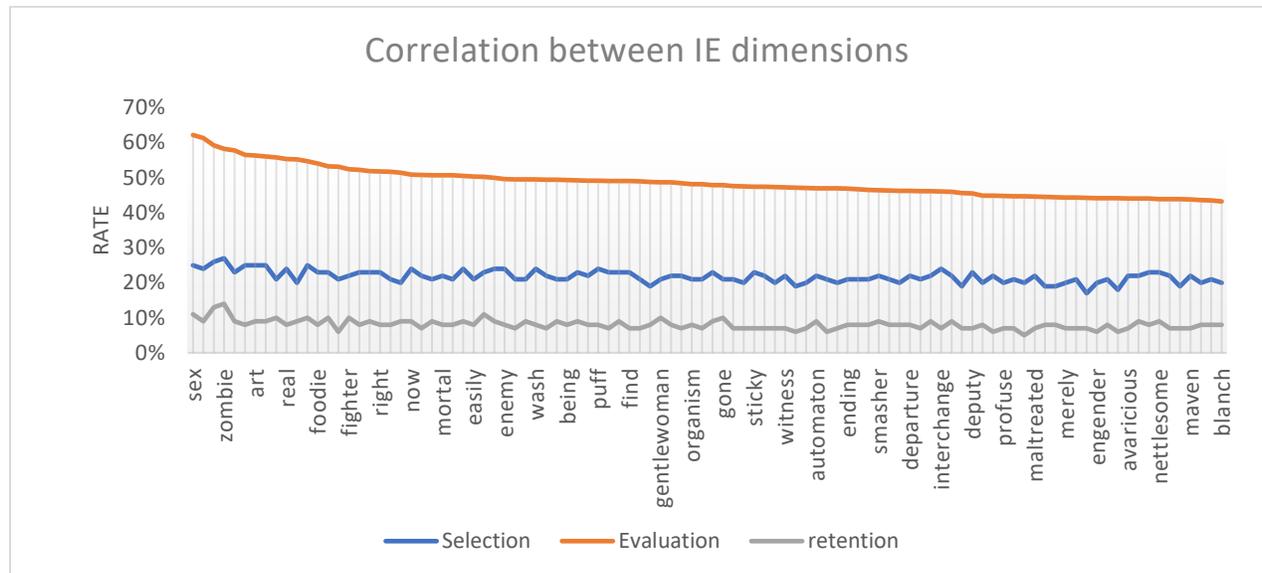

**Figure 8. Correlations among IE Dimensions of Selected Words**

## 4    Discussion

### 4.1    Hypothesis Testing Findings

The results provide evidence that use of different words to express the same information yield different evaluation, selection, and retention rates. Moreover, they strongly indicate that the rates of evaluation, selection, and retention, which are used to measure the IE dimensions of  perception, participation, and perseverance, respectively, are signficantly positively correlated. Together these findings suggest that IE is a driver for information use as operationalized by perception (evaluation), participation (retrieval or selection rate), and perseverance (retention and integration or memorization). By providing support that IE is an important research construct, these findings reveal that its enhancement may lead to a desired outcome in the form of positive affective, behavioral, and cognitive responses.

## 4.2    Research Questions

The results of the hypothesis testing answered Research Question 1—What are the dimensions and determinants of IE?—by confirming that perception, participation, and perseverance are the most signficant determinants of IE and can be measured by determining rates of evaluation, selection, and retention, respectively. The variance in word IE scores revealed that IE is driven, if not determined, *solely* by how information is expressed by word choice *to the exclusion* of other factors. IE is therefore distinct from user UE and UX because it depends *only* on the information itself and is not dependent on user, task, or system variables, e.g., user age, used for work or school, or IS features, respectively. The results also revealed that variations in word choice, even when using synonyms, significantly impact levels of perception, participation, and perseverance. Therefore, the results provide evidence that Research Question 2—Can changes in word choice and the phrasing of information impact IE?—can be answered in the affirmative.

## 5    Conclusion

This study contributes to the literature by identifying the dimensions of IE; their relationships among each other; and their significance in word choice and, ultimately, decision making among producers of digital textual information. The findings provide evidence that perception, participation, and perseverance, the three dimensions of IE, as positively correlated and determined by word choice. These findings in turn strongly indicate that language use, specifically word choice, has a significant impact on how consumers of digital text perceive and respond to the information it conveys.

Moreover, and perhaps most significantly, the findings reveal that the nature of and extent of a user's interactions with digital textual information (i.e., participation, and perseverance) very likely depend *solely* on of the digital textual information itself and is not influenced by the IS used. Therefore, the same information presented on different systems will yield the same results in terms of perception, participation, and perseverance.

These findings have important implications for content producers seeking to maximize IE among readers of their digital textual information. Rather than focusing on optimizing the way in which different systems present the same information, they should seek to optimize the choice of words and the order of their presentation (i.e., syntax and phrasing) in their text and focus relatively less on IS factors such as physical "catchiness." By determining how the words should be chosen and presented and not how they look on an IS, future research can continue exploring how IE can be increased to the greatest extent possible. The findings can provide guidelines for content producers seeking to increase their target readership's interaction with their digital textual information to the greatest extent possible, promoting enhanced communication with readers and decision making by readers. The findings can also aid in the development of engagement strategies targeted toward different populations and evaluate how engagement varies across different populations and contexts.